\documentclass[letterpaper, conference, 10pt,twocolumn ]{IEEEtran}
\pdfoutput=1
\let\chapter\section
\usepackage[ruled,vlined,linesnumbered]{algorithm2e}
\usepackage{graphicx}
\usepackage{rotating}
\usepackage{floatrow}
\usepackage{amsfonts}
\usepackage{amsmath,amssymb,mathrsfs}
\usepackage[singlelinecheck=off]{caption}
\usepackage{array}
\usepackage{flafter}
\usepackage{subfigure}
\usepackage{color}
\usepackage[numbers,sort&compress]{natbib}
\usepackage{bm} 
\usepackage{subfigure}
\usepackage{multirow}
\usepackage{epstopdf}
\usepackage{afterpage}

\newcommand{\txcm}[1]{\textcolor{magenta}{#1}}
\newcommand{\txcb}[1]{\textcolor{blue}{#1}}

\newcommand{\pbold}[1]{\textbf{#1:} }

\IEEEoverridecommandlockouts

\begin{document}

\title{RFM-SLAM: Exploiting Relative Feature Measurements to Separate Orientation and Position Estimation in SLAM}

\author{\IEEEauthorblockN{Saurav Agarwal}
	\and
	\IEEEauthorblockN{Vikram Shree}
	\and
	\IEEEauthorblockN{Suman Chakravorty}
	\thanks{Saurav Agarwal (\texttt{sauravag@tamu.edu}) and Suman Chakravorty (\texttt{schakrav@tamu.edu}) are with the Department of Aerospace Engineering, Texas A\&M University, College Station, TX 77840, USA. Vikram Shree (\texttt{vikshree@iitk.ac.in}) is with the Department of Aerospace Engineering, Indian Institute of Technology, Kanpur, India.}}

\maketitle

\begin{abstract}

The SLAM problem is known to have a special property that when robot orientation is known, estimating the history of robot poses and feature locations can be posed as a standard linear least squares problem. In this work, we develop a SLAM framework that uses relative feature-to-feature measurements to exploit this structural property of SLAM.
Relative feature measurements are used to pose a linear estimation problem for pose-to-pose orientation constraints. This is followed by solving an iterative non-linear on-manifold optimization problem to compute the maximum likelihood estimate for robot orientation given relative rotation constraints. Once the robot orientation is computed, we solve a linear problem for robot position and map estimation. Our approach reduces the computational burden of non-linear optimization by posing a smaller optimization problem as compared to standard graph-based methods for feature-based SLAM. 
Further, empirical results show our method avoids catastrophic failures that arise in existing methods due to using odometery as an initial guess for non-linear optimization, while its accuracy degrades gracefully as sensor noise is increased. We demonstrate our method through extensive simulations and comparisons with an existing state-of-the-art solver. 

Keywords: SLAM, graph-based SLAM, non-linear optimization, relative measurements
\end{abstract}

\section{Introduction}

Relative measurements \cite{newman-projectionfilter, durrant-whyte-acc97, Sibley-RSS-09, martinelli2007relative, pradalier-sekhavat} allow a robot to exploit structural properties of the environment, e.g., relative displacement from one landmark to another is independent of how a robot moves in a static world given a particular frame of reference.
Taking note of this property, we present a 2D SLAM approach in which range bearing measurements are transformed into relative displacements between features. In our method, relative orientation constraints between poses are formulated using translation and rotation invariant structural properties. This allows our method to exploit the separable structure of SLAM \cite{Khosoussi-RSS-15, Carlone-LAGO, carlone-censi-tro14}, i.e., robot heading estimation is separated from the estimation of past robot positions and feature locations. Using relative orientations between the set of robot poses, our method solves a non-linear optimization problem over the set of robot orientations
following which we solve a linear least squares problem for position (robot trajectory and map). We call this method Relative Feature Measurements-based Simultaneous Localization and Mapping (RFM-SLAM). 

\begin{figure}[ht!]
	\centering
	\subfigure[\txcb{RFM-SLAM} estimate for robot trajectory in one of our simulations. RMS position error is $1.88$m.]{\includegraphics[width=1.55in]{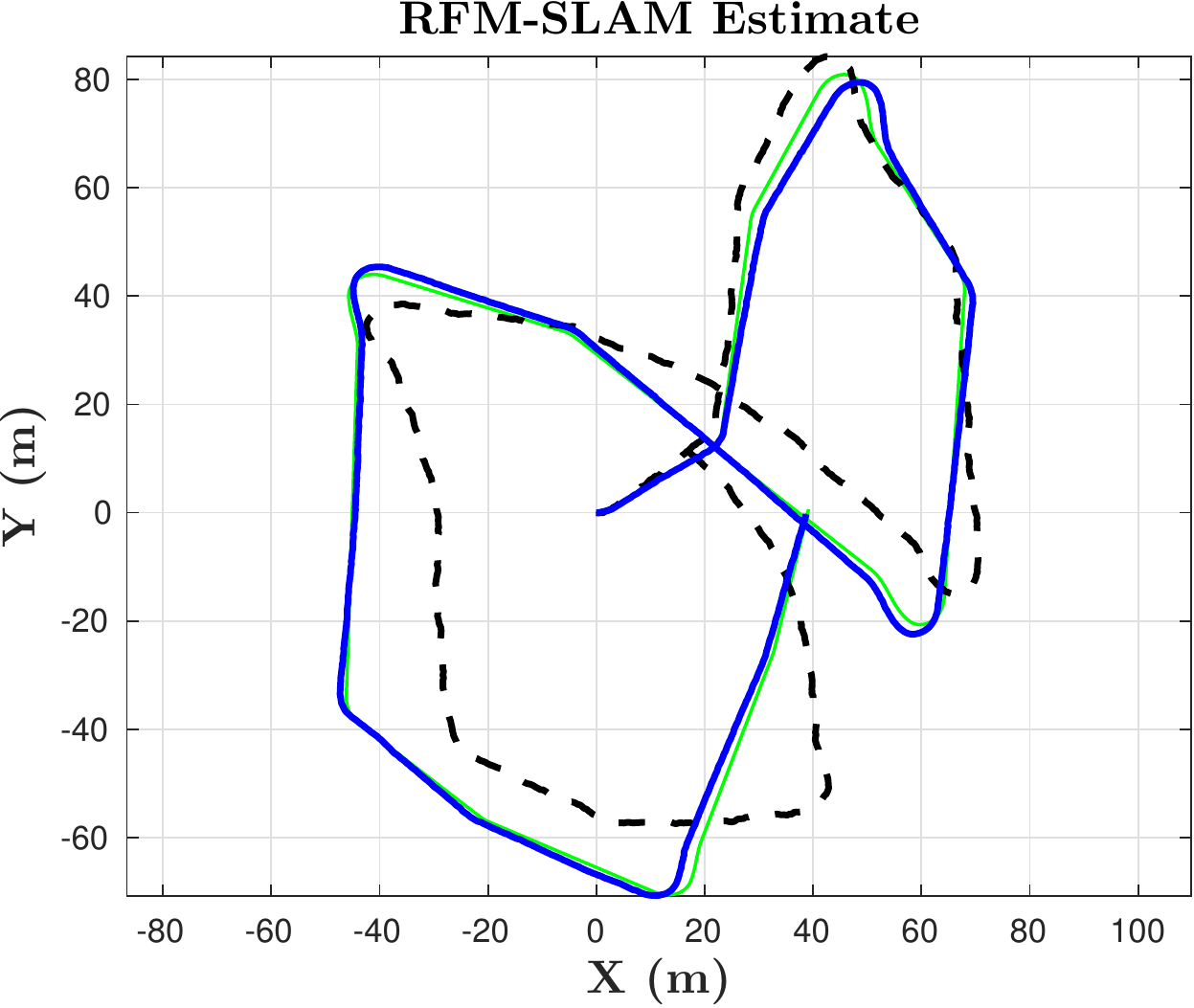}\label{subfig:rfmslam-mapl1-intro}}
	 \hspace{0.1in}	
	\subfigure[\txcm{GTSAM} estimate for robot trajectory in the same run, catastrophic failure due to bad initial guess.]{\includegraphics[width=1.55in]{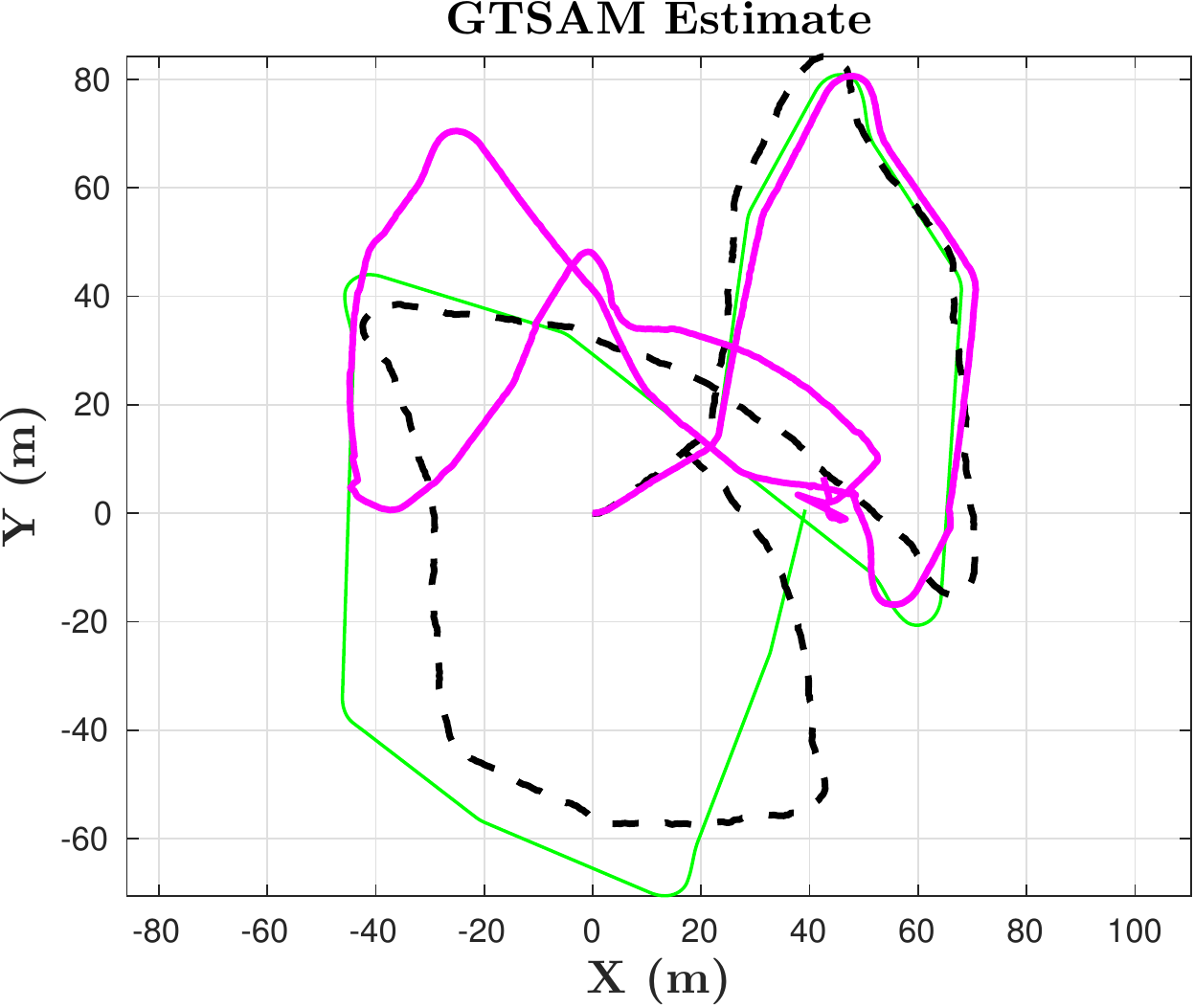}\label{subfig:gtsamam-mapl1-intro}}
	\caption{Simulation results for map M1 with $\approx 1000$ nodes for \txcb{RFM-SLAM} and \txcm{GTSAM} given identical data. The \textcolor{green}{true trajectory and landmarks} are in green, \textbf{odometery} is in black, \txcb{RFM-SLAM} estimates are shown in blue and \txcm{GTSAM} estimates in magenta. Feature plots are omitted for the sake of clarity.}
	\label{fig:rfm-gtsam-comparison-M1}
\end{figure}	

\begin{figure}[ht!]
	\subfigure[\txcb{RFM-SLAM} estimate for robot trajectory in one of our simulations. RMS position error is $1.44$m.]{\includegraphics[width=1.55in]{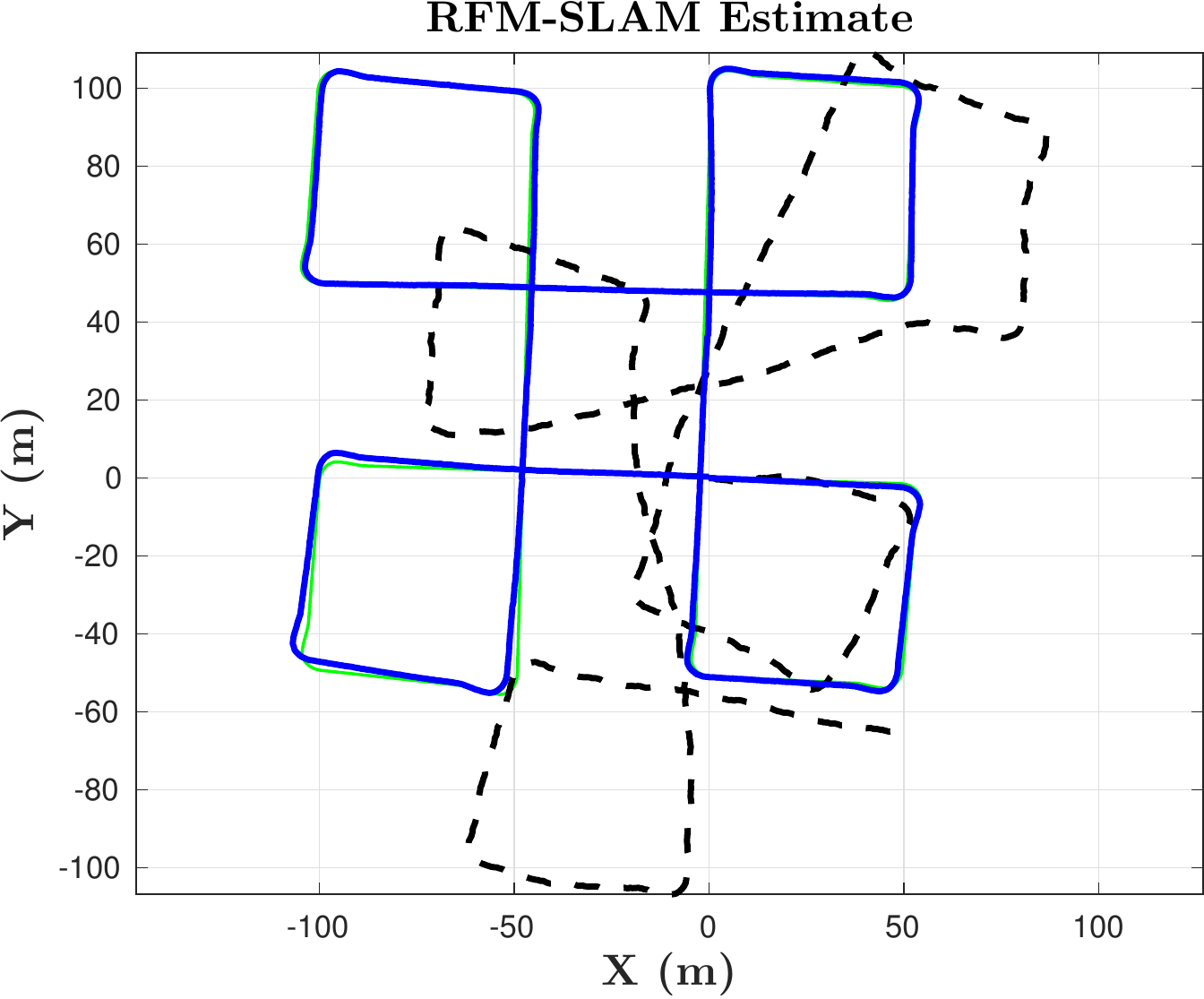}\label{subfig:rfmslam-maps4-intro}}		
	\hspace{0.1in}
	\subfigure[\txcm{GTSAM} estimate for robot trajectory for the same run, catastrophic failure due to bad initial guess.]{\includegraphics[width=1.55in]{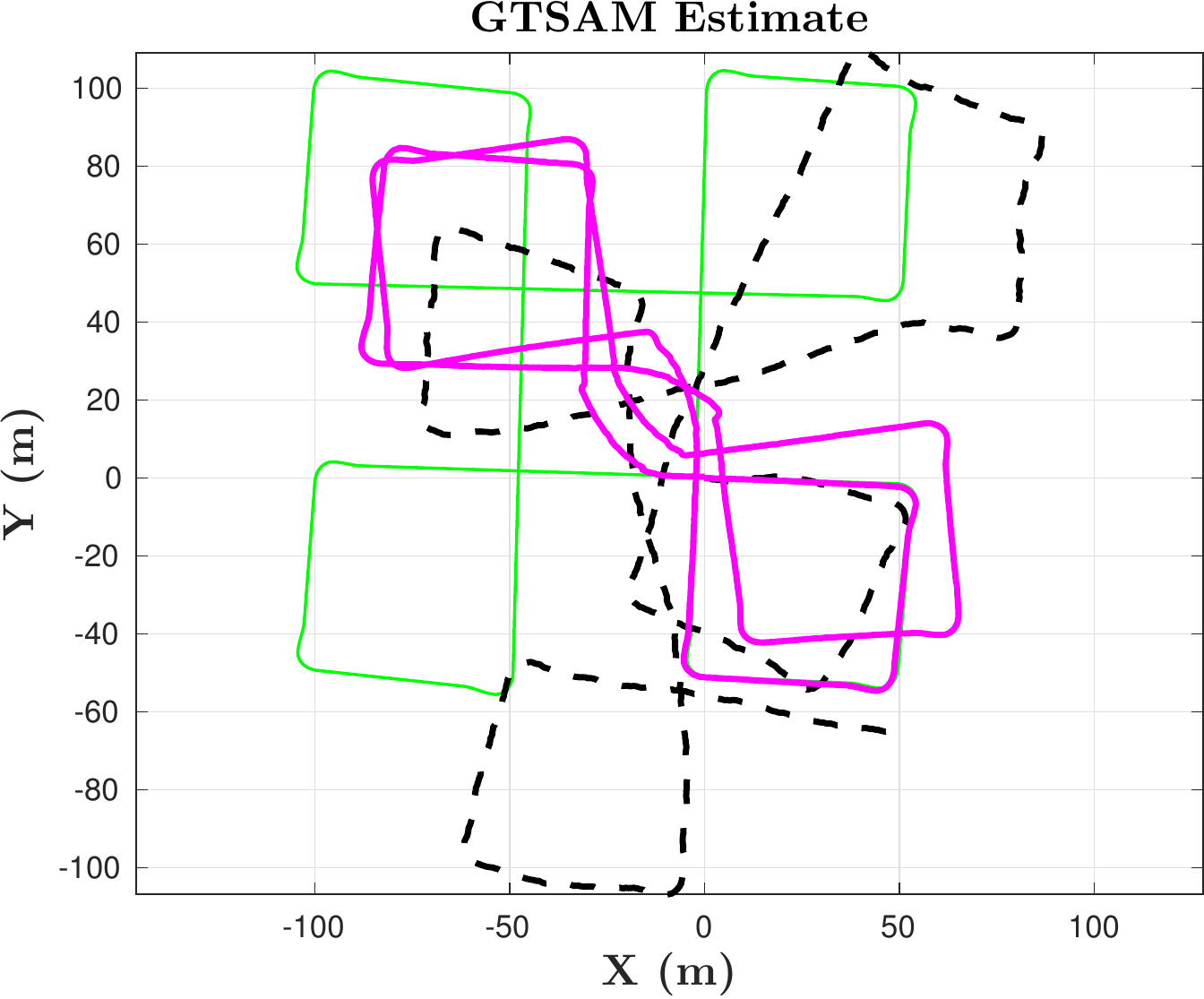}\label{subfig:gtsam-maps4-intro}}	
	\caption{Simulation results for map M2 with $\approx 2000$ nodes for \txcb{RFM-SLAM} and \txcm{GTSAM} given identical data.}
	\label{fig:rfm-gtsam-comparison-M2}
\end{figure}

Figures \ref{fig:rfm-gtsam-comparison-M1} and \ref{fig:rfm-gtsam-comparison-M2} shows a comparison between RFM-SLAM and GTSAM \cite{dellaert2012factor} for two maps. When the simulated data is input to GTSAM, the odometery based initial guess results in catastrophic failure for both maps as GTSAM gets stuck in a local minima whereas RFM-SLAM recovers the robot trajectory and map given identical data. 
The major contributions of this paper can be summed up as follows:

\begin{enumerate}
	\item RFM-SLAM reduces computational complexity of the optimization problem, i.e., if there are $N$ poses where each pose $\mathbf{x}_k= [\mathbf{p}_k, \boldsymbol{\theta}_k]^T$ and $L$ landmarks then we solve for $N$ variables as opposed to $3N + 2L$ in existing methods (for the planar SLAM problem).
	\item By separating orientation estimation and formulating the robot and landmark position estimation as a linear least squares problem, no initial guess is required for the positions. Further, we show through empirical results that as odometery noise increases, our method avoids catastrophic failures which may occur in non-linear optimization-based methods due to reliance on odometery-based initial guess.
\end{enumerate}

We now proceed to discuss relevant related work. In Section \ref{sec:splam-problem} we state our problem and preliminaries, subsequently in Section \ref{sec:method-splam} we present our approach wherein some mathematical details are relegated to Appendix \ref{appx:llsq-rotations}. Results are discussed in Section \ref{sec:plum1-results} followed by conclusions.

\section{Related Work}\label{sec:litsurvey}

The initial work of \cite{SmithCheesemanISRR} introduced filtering as a tool to tackle the SLAM problem. Several later works \cite{martinelli2007relative,durrant-whyte-acc97, newman-projectionfilter} proposed to exploit relative feature measurements in a filtering-based approach. In \cite{durrant-whyte-acc97} the correlations between relative measurements from common landmarks are not considered which leads to a sub-optimal estimate. In \cite{newman-projectionfilter} only relative distances are estimated which neglects the information provided by the direction component of relative measurements. The method of \cite{martinelli2007relative} exploits the shift and rotation invariance of map structure but cannot consistently incorporate long range correlations and is thus unable to close loops. In comparison to aforementioned methods \cite{martinelli2007relative,durrant-whyte-acc97, newman-projectionfilter} our formulation takes into account both; correlations between relative measurements from common landmarks; and long range correlations between relative measurements in the global frame. This allows RFM-SLAM to form consistent estimates and close large loops.
The method of \cite{Wang-dslam} exploits relative feature measurements to decouple map estimation from robot localization in an Extended Information Filter-based formulation, while maintaining long range correlations. Compared to \cite{Wang-dslam, martinelli2007relative,durrant-whyte-acc97, newman-projectionfilter} we exploit relative measurements to decouple robot orientation estimation from map and robot position. Further RFM-SLAM does not maintain a recursive estimate over the map or robot state, it falls into the category of methods that solve the full SLAM problem.

The seminal contribution of \cite{lu-milios-1997} introduced a non-linear optimization based approach to solving the full SLAM problem wherein robot poses are treated as nodes of a graph and constraints as edges. In \cite{thrun-graphslam} the authors extended graph-based SLAM to feature mapping and several others \cite{Dellaert-sqrtslam, Konolige-2010, folkesson-icra2004, mahon-tro2008, Kaess08tro, Kaess-isam2, Kaess2011} made significant contributions to extend the initial work of \cite{lu-milios-1997}. A key limitation of non-linear iterative optimization methods is that an initial guess is required to bootstrap the solver and this guess is usually provided by odometery. However, it is well known that odometery error grows unbounded and is often unreliable.
This reliance on odometery for initial guess makes non-linear optimization methods susceptible to getting trapped in local minima often resulting in arbitrarily bad solutions \cite{carlone-censi-tro14,carlone-tron-icra15}  (sometimes referred to as catastrophic failures, see Figs. \ref{fig:rfm-gtsam-comparison-M1} and \ref{fig:rfm-gtsam-comparison-M2}).
Recent works \cite{Carlone-LAGO, carlone-censi-tro14, Khosoussi-RSS-15} have analyzed structural properties of SLAM with the aim of decoupling non-linearities that arise due to orientation. The works of \cite{Carlone-LAGO, carlone-censi-tro14} provided several important insights, demonstrating that estimating orientation as the first step and using these estimates to initialize pose graph optimization results in a robust solution.  In \cite{boumal-cdc13} a general on-manifold optimization based method is developed to estimate orientations from noisy relative measurements corrupted by outliers. In relation to \cite{boumal-cdc13}, our orientation estimation method (Section \ref{subsec:heading-estimation-manopt}) is only concerned with measurement data corrupted by zero-mean Gaussian noise similar to \cite{Carlone-LAGO, Khosoussi-RSS-15}. We direct the reader to \cite{carlone-tron-icra15} for a recent survey of 3D rotation estimation techniques. The works of \cite{Carlone-LAGO, Khosoussi-RSS-15} are closely related to ours, hence we proceed to discuss these in greater detail.

Linear Approximation for pose Graph Optimization (LAGO) \cite{Carlone-LAGO} is a method for planar relative pose graph SLAM that separates robot orientation and position estimation into two successive linear problems with the key benefit of a reduced risk of convergence to local minima and provides a robust initial guess for iterative optimization. The LAGO formulation does not deal with feature-based measurements and cannot be extended to 3D. In contrast, RFM-SLAM is designed for feature-based SLAM and majority of the algorithm presented in this paper ports directly to the 3D domain (see discussion in Section \ref{subsec:method-remarks}). LAGO develops a closed form approach (regularization) to solve the angle wrap-around problem that relies on rounding-off noisy relative orientation measurements. This technique may degrade rapidly once sensor accuracy reduces beyond a certain threshold (\cite{Carlone-LAGO}, Section 6). In contrast, RFM-SLAM does not invoke any such approximation as it computes the maximum likelihood estimate for the orientations via an on-manifold optimization. In this regard, compared to LAGO, our approach trades computational speed, for accuracy and reliability in the orientation estimation phase.
In \cite{Khosoussi-RSS-15}, the authors develop a modified Variable Projection (VP) technique for non-linear optimization that exploits the separation of position and orientation in SLAM and runs faster than the standard Gauss Newton algorithm. The method of \cite{Khosoussi-RSS-15} solves for orientation and position successively in an iterative manner as opposed to RFM-SLAM wherein iterative non-linear optimization is only applied to orientation estimation. The method of \cite{Khosoussi-RSS-15} may get trapped in local minima and in few instances may not converge to a solution (\cite{Khosoussi-RSS-15}, Section 5) as it relies on odometery for the initial guess which may be arbitrarily bad. Our empirical observations indicate that as sensor noise is increased, RFM-SLAM performance degrades gracefully and we do not observe catastrophic failures (see Table \ref{table:rmse-50runs-rfmslam}). 
\section{Preliminaries and Problem}\label{sec:splam-problem}

Let $ x_{k} \in \mathbb{X} $, $ u_{k} \in \mathbb{U} $, and $ z_{k} \in \mathbb{Z}$ represent the system state, control input, and observation at time step $ k $ respectively, where $\mathbb{X}, \mathbb{U}, \mathbb{Z}$ denote the state, control, and observation spaces respectively. The measurement model $h$ is denoted as $ z_{k} =h(x_{k}) + v_{k}$, where $ v_{k} \sim \mathcal{N}(0,\mathbf{R}_k)$ 
is zero-mean Gaussian measurement noise. The map (unknown at $t_0$) is a set of landmarks (features) distributed throughout the environment. We define the $j$-th landmark as $l_j$ and $\hat{l}_j$ as the estimate of $l_j$. The observation for landmark $l_j$ at time $t_k$ is denoted by $z^{j}_{k} \in z_{k}$. The inverse measurement model is denoted by $g$ such that for a given measurement $z^{j}_k$ and the state $x_k$ at which it was made, $g$ computes the landmark location $l_j = g(x_k, z^{j}_k)$. The state evolution model $f$ is denoted as  $x_{k+1}=f(x_{k},u_{k}) + w_{k}$ where $w_{k} \sim \mathcal{N}(0,\mathbf{Q}_k)$ is zero-mean Gaussian process noise.

\begin{figure*}
	\subfigure[Robot making local relative measurements.]{\includegraphics[height=1.1in]{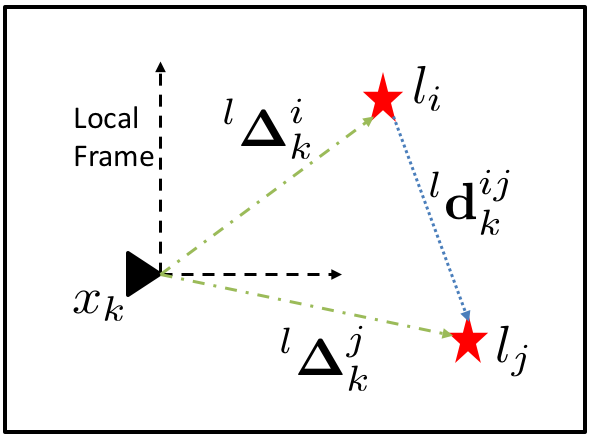}\label{subfig:rel-obs-simple}}
	\hspace{0.25in}
	\subfigure[Robot observes same features from two different poses forming a relative rotation constraint.]{\includegraphics[height=1.1in]{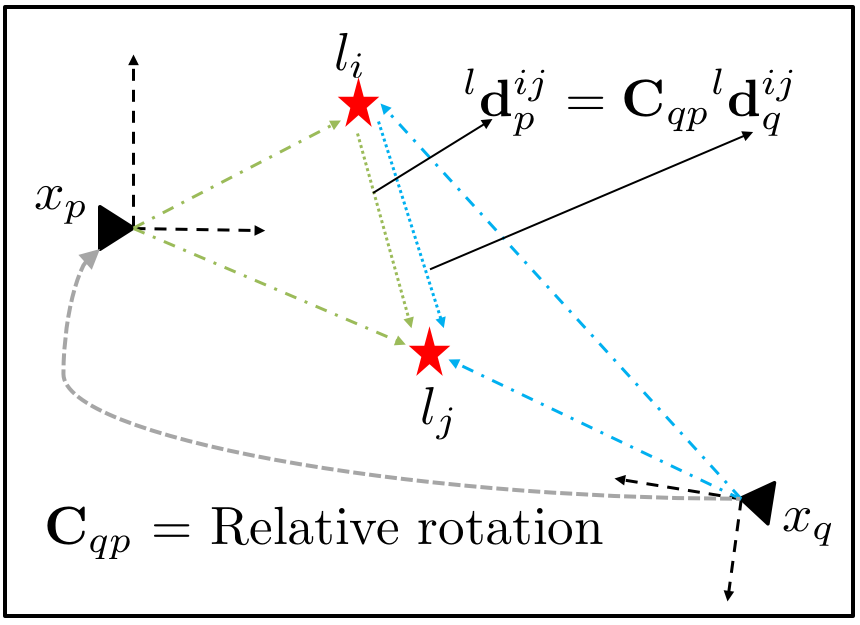}\label{subfig:rel-orientation-2-poses}}
	\hspace{0.25in}
	\subfigure[Tranformation of local robot to feature relative measurements to the global frame.]{\includegraphics[height=1.1in]{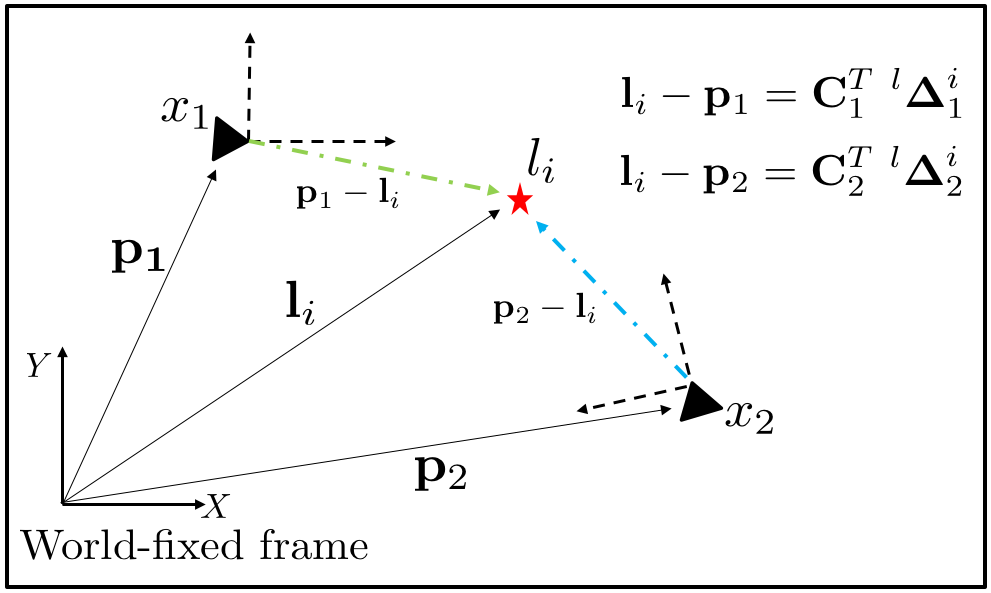}\label{subfig:global-position}}
	\caption{(a) A robot making observations to two features $l_i$ and $l_j$ at time $t_k$, the range bearing measurements allow the robot to compute the relative positions ${^l}\Delta^i_k$ and ${^l}\Delta^j_k$ of the features in its local frame which are then transformed to a relative displacement measurement ${^l}d^{ij}_k$ between the two features. (b) A robot making observations to two features from poses $x_p$ (green arrows) and $x_q$ (blue arrows). Seeing the same two features forms a rotation constraint $C_{qp}$ between these poses. (c) A robot sees the same landmark from two poses, the transformation of local relative measurements to the global frame is used in Section \ref{subsec:global-estimation} to solve for robot and feature positions.}
	\label{fig:rfmslam-mapping-cartoon}
\end{figure*}

We define ${^l}\mathbf{d}^{ij}_k$ to be the relative feature measurement, from feature $l_i$ to $l_j$ in the local frame of the robot at time $t_k$. In our framework, a relative feature measurement is an estimate of the displacement vector from one feature to another (Fig. \ref{subfig:rel-obs-simple}). The local relative measurement is computed as ${^l}\mathbf{d}^{ij}_k = {^l}\boldsymbol{\Delta}^j_k - {^l}\boldsymbol{\Delta}^i_k$, where ${^l}\boldsymbol{\Delta}^i_k, {^l}\boldsymbol{\Delta}^j_k$ are relative positions of features $l_i$ and $l_j$ respectively with respect to the robot in its local frame. Thus it is linear in positions of the two features in the local frame.
Let $\mathbf{C}(\boldsymbol{\theta}_k)$ denote the Direction Cosine Matrix (DCM) of the robot orientation at state $x_k$. $\mathbf{C}$ is a function of the robot orientation parameter $\boldsymbol{\theta}_k$ (e.g., Euler angles, Quaternions etc.). A local measurement in the robot frame can be projected into the world (global) frame as

\begin{equation}
 \mathbf{C}(\boldsymbol{\theta}_k)^T ~ {^l}\boldsymbol{\Delta}^i_k = {^w}\boldsymbol{\Delta}^i_k = \mathbf{l}_i - \mathbf{p}_k,
\end{equation}

\noindent
where $\mathbf{l}_i$ and $\mathbf{p}_k$ are the feature and robot positions in the world frame. Thus, it is the transformation of local measurements to the global frame that introduces non-linearity due to the trigonometric functions of orientation. If heading $\boldsymbol{\theta}^{*}$ is known, define ${^l}\boldsymbol{\Delta}$ to be the vector of all local feature position measurements and let $[\mathbf{p}^T ~ \mathbf{l}^T]^T$  be the vector of all robot and feature positions in the world frame, then we have the following standard linear estimation problem in position

\begin{equation}\label{eq:local2global-positions}
\mathbf{C}(\boldsymbol{\theta}^{*})^T ~ {^l}\boldsymbol{\Delta} = \mathbf{A}' \begin{bmatrix}
\mathbf{p} \\
\mathbf{l}
\end{bmatrix},
\end{equation}

where $\mathbf{A}'$ is a matrix composed of elements in the set $\{-1,0,1\}$. However, direct heading estimates may not be readily available due to which we need to estimate the robot heading. In the proceeding section we develop the RFM-SLAM algorithm and describe our heading and position estimation method in detail. It is assumed that relative orientation measurements are independent and the front-end is given, the focus of this paper is on the back-end estimation problem.

\section{Methodology}\label{sec:method-splam}

The key steps in RFM-SLAM are as follows:

\begin{enumerate}
	\item Transform range bearing observations from robot to features into relative position measurements in the robot's local frame at each pose, then calculate feature-to-feature displacements vectors (Section \ref{subsec:method-rel-landmarks}).
	\item Compute the relative rotation constraints for poses that either are connected by proprioceptive odometery or view identical pairs of landmarks or both (Section \ref{subsec:heading-estimation-llsq}).
	\item Compute the Maximum Likelihood Estimate (MLE) for the robot orientation given constraints computed in the previous step (Section \ref{subsec:heading-estimation-manopt}). 
	\item Solve the global linear estimation problem over robot and feature positions (Section \ref{subsec:global-estimation}).
\end{enumerate}

\subsection{Relative Feature Displacement Estimation}\label{subsec:method-rel-landmarks}

Figure \ref{fig:rfmslam-mapping-cartoon} depicts our proposed feature mapping process. 
At time $t_k$ let the robot make range bearing measurements $z^{i}_k$ and $z^{j}_k$ to landmarks $l_i$ and $l_j$ respectively. Using the inverse measurement model $g$ (Section \ref{sec:splam-problem}), we have the position of $l_i$ in robot's local frame as ${^l}\boldsymbol{\Delta}^i_k = {^l}\mathbf{g}_{\Delta}(\mathbf{z}^{i}_k)$ and ${^l}\boldsymbol{\Delta}^j_k = {^l}\mathbf{g}_{\Delta}(\mathbf{z}^{j}_k)$ for $l_j$. Thus, we can compute the displacement vector from $l_i$ to $l_j$ in the robot frame as

\begin{flalign}\label{eq:rel-obs}
\nonumber
{^l}\mathbf{d}^{ij}_k  = {^l}\boldsymbol{\Delta}^j_k - {^l}\boldsymbol{\Delta}^i_k & = {^l}\mathbf{g}_{\Delta}(\mathbf{z}^{j}_k) - {^l}\mathbf{g}_{\Delta}(\mathbf{z}^{j}_k) \\
& = {^l}\mathbf{g}_{d}(\mathbf{z}^{j}_k,\mathbf{z}^{i}_k).
\end{flalign}

${^l}\mathbf{g}_{d}(\mathbf{z}^{j}_k,\mathbf{z}^{i}_k)$ is the relative measurement from $l_i$ to $l_j$ in the robot's frame, which is independent of robot position and orientation. Figure \ref{subfig:rel-obs-simple} shows a simple depiction of a robot making a relative position measurement between two features. 
Let $L_{k} = \{l_{k_1},l_{k_2},\dots,l_{k_n}\}$ be the set of landmarks visible at time $t_k$ and $z_k = \{z^{l_{k_1}}_k,z^{l_{k_2}}_k,\dots,z^{l_{k_n}}_k\}$ be the set of range bearing observations to the same. Let $D_k = \{d^{l_{k_1} l_{k_2}}, d^{l_{k_1} l_{k_3}},\dots,  d^{l_{k_{n-1}} l_{k_n}}\}$ be the set of relative observations between these features, where $|D_k| = {|z_k| \choose 2}$. Hence, the vector of local relative measurements is as follows

\begin{equation}\label{eq:local-relative-meas-func}
{^l}\mathbf{\hat{d}}_k = {^l}\mathbf{g}_{d}(\mathbf{z}_k)=
\begin{bmatrix}
{^l}\mathbf{g}_{d}(\mathbf{z}^{l_{k_1}}_k,\mathbf{z}^{l_{k_2}}_k) \\
{^l}\mathbf{g}_{d}(\mathbf{z}^{l_{k_1}}_k,\mathbf{z}^{l_{k_3}}_k)\\
\vdots \\
{^l}\mathbf{g}_{d}(\mathbf{z}^{l_{k_{n-1}}}_k,\mathbf{z}^{l_{k_n}}_k)
\end{bmatrix}.
\end{equation}

To estimate the error covariance of the relative measurement in Eq. \ref{eq:local-relative-meas-func} we linearize ${^l}\mathbf{g}_{d}(\mathbf{z}_k)$. Let $\bar{\nabla}{^l}\mathbf{g}_{d}|_{\mathbf{z}_k}$ to be the Jacobian of the local relative measurement function in Eq. \ref{eq:local-relative-meas-func} and let $\mathbf{R}_{\mathbf{z}_k}= diag([\mathbf{R}_{l_{k_1}}, ~\mathbf{R}_{l_{k_2}}, \dots ])$ be the noise covariance of $\mathbf{z}_k$, where $\mathbf{R}_{l_{k_i}}$ is the noise covariance of robot's range bearing measurement to feature $l_{k_i}$. It is important to note that though measurements to each feature are independent, the set of relative feature measurements is \textit{correlated}. This can be attributed to the correlations between relative measurements from common landmarks (see Eq. \ref{eq:local-relative-meas-func}). Finally, we have ${^l}\mathbf{\hat{d}}_k \sim \mathcal{N}({^l}\mathbf{d}_k, {^l}\mathbf{R}_{\mathbf{d}_k} = \bar{\nabla}{^l}\mathbf{g}_{d}|_{\mathbf{z}_k} \mathbf{R}_{\mathbf{z}_k} \bar{\nabla}{^l}\mathbf{g}^T_{d}|_{\mathbf{z}_k})$.

\subsection{Heading Estimation}

We now proceed to develop a two-part heading estimation technique. First, we recognize the fact that relative feature measurements-based constraints on the rotation between two poses are linear in the elements of relative orientation Direction Cosine Matrix (DCM). Thus we propose a linear least squares formulation to estimate the relative rotation between poses. 
The second step is described in Section \ref{subsec:heading-estimation-manopt} where we apply an on-manifold optimization approach to solve the general non-linear heading estimation problem at loop closure given relative orientation estimates.

\subsubsection{Linear Relative Rotation Estimation}\label{subsec:heading-estimation-llsq}

Let $\mathbf{C}_{qp}$ be the relative rotation matrix between two poses $x_p,x_q$ such that $\mathbf{v}_p = \mathbf{C}_{qp} \mathbf{v}_q$, i.e, the vector $\mathbf{v}$ in frame $q$ can be transformed through $\mathbf{C}_{qp}$ to frame $p$. We know that $\mathbf{C}_{qp} = \mathbf{C}_{p} \mathbf{C}^T_{q}$ where $\mathbf{C}_{p}, \mathbf{C}_{q} \in \text{SO}(2)$. Let there be two landmarks $l_i, l_j$ visible from poses $x_p, x_q$. Let ${^l}\mathbf{d}^{ij}_{p}, {^l}\mathbf{d}^{ij}_{q}$ be the vectors from $l_i$ to $l_j$ in the local frames at each pose. Then we have a constraint ${^l}\mathbf{d}^{ij}_{p} - {^l}\mathbf{C}_{qp} \mathbf{d}^{ij}_{q} = \mathbf{0}$ for every pair of landmarks $(l_i,l_j)$ visible from $x_p$ and $x_q$. Let $\mathbf{c}_{qp} \in \mathbb{R}^2$ be the vector of parameters for $\mathbf{C}_{qp}$ (see Eq. \ref{eq:2d-dcm-parameters}, Appendix \ref{appx:llsq-rotations}).
As a robot moves, it makes two types of noisy observations:
\begin{enumerate}

\item Proprioceptive odometery measurements $\widehat{\delta \theta}_{odo} \sim \mathcal{N}(\delta \theta, \sigma^2_{odo})$ provide a direct estimate of the relative rotation $\delta \theta$ between successive poses $x_p$ and $x_{q=p+1}$, where $\sigma^2_{odo}$ is the measurement noise variance. Thus the vector $\mathbf{\hat{c}}_{qp,odo} = [cos(\widehat{\delta \theta}_{odo}), sin(\widehat{\delta \theta}_{odo})]^T$.

\item Relative feature measurements to common landmarks from two poses provide a relative orientation constraint. Define $D_{pq} = D_p \cap D_q$ to be the set of common relative measurements made from poses $x_p$ and $x_q$, and 
${^l}\mathbf{\hat{d}'}_{p} \subseteq {^l}\mathbf{\hat{d}}_{p}, {^l}\mathbf{\hat{d}'}_{q} \subseteq {^l}\mathbf{\hat{d}}_{q}$ be the respective local measurements in the set $D_{pq}$ with error covariances ${^l}\mathbf{R}'_{\mathbf{d}_{p}}, {^l}\mathbf{R}'_{\mathbf{d}_{q}}$ respectively. We have the following linear problem for the relative rotation parameter vector $\mathbf{c}_{qp}$, 

\begin{equation}\label{eq:rot-constraint-rel-feats}
{^l}\mathbf{\hat{d}'}_{p} = \mathbf{B}'_{qp} \mathbf{c}_{qp} + \mathbf{v}_{\mathbf{d}_{pq}},
\end{equation}

where $\mathbf{B}'_{qp} ({^l}\mathbf{\hat{d}'}_{q})$ (see Eq. \ref{eq:det-rot-constraint}, Appendix \ref{appx:llsq-rotations}) is a matrix function of the relative measurements from pose $x_q$ and $\mathbf{v}_{\mathbf{d}_{pq}} \sim \mathcal{N}(\mathbf{0}, \mathbf{R}_{\mathbf{d}_{pq}})$ is a zero-mean Gaussian measurement noise. 
The error covariance in this measurement is approximated as $\mathbf{R}_{\mathbf{d}_{pq}} = {^l}\mathbf{R}'_{\mathbf{d}_{p}} + \mathbf{\hat{C}}_{qp, init}{^l}\mathbf{R}'_{\mathbf{d}_{q}}\mathbf{\hat{C}}^T_{qp, init}$.

\end{enumerate}

For successive poses, $\mathbf{\hat{C}}_{qp, init} = \mathbf{\hat{C}}_{qp, odo}$, i.e., the relative rotation estimate from proprioceptive odometery. Between successive poses, all feature constraints in the form of Eq. \ref{eq:rot-constraint-rel-feats} can be stacked along with proprioceptive odometery measurements which gives us the following linear problem

\begin{flalign}\label{eq:llsq-rot-gen}
\nonumber
\begin{bmatrix}
\mathbf{\hat{c}}_{qp,odo}\\
{^l}\mathbf{\hat{d}}_{p} \\
\end{bmatrix}
 & = \begin{bmatrix}
\mathbf{I} \\
\mathbf{B}'_{qp}
\end{bmatrix}
\mathbf{c} +
\begin{bmatrix}
\mathbf{v}_{\mathbf{c}_{qp,odo}} \\
\mathbf{v}_{\mathbf{d}_{pq}}
\end{bmatrix}\\
 & = \mathbf{B}_{qp} \mathbf{c} + \mathbf{v}_{\mathbf{c}_{pq}}
\end{flalign}

where $\mathbf{v}_{\mathbf{c}_{pq}} \sim \mathcal{N}(\mathbf{0}, \mathbf{R}_{\mathbf{c}_{qp}})$, and $\mathbf{R}_{\mathbf{c}_{qp}} = diag([\mathbf{R}_{\mathbf{c}_{qp,odo}},\mathbf{R}_{\mathbf{d}_{pq}} ])$. Equation \ref{eq:llsq-rot-gen} can be rewritten as 

\begin{equation}\label{eq:heading-linear-problem}
\mathbf{z}_{\mathbf{c}_{qp}} = \mathbf{B}_{qp}\mathbf{c}_{qp} + \mathbf{v}_{\mathbf{c}_{qp}}.
\end{equation}

Dropping the pose subscript for clarity, we can compute the estimate $\hat{\mathbf{c}} = (\mathbf{B}^T \mathbf{R}^{-1}_{\mathbf{c}} \mathbf{B})^{-1}\mathbf{B}^T \mathbf{R}^{-1}_{\mathbf{c}} \mathbf{z}_{\mathbf{c}}$ and its error covariance $\boldsymbol{\Sigma}_{\mathbf{c}} = (\mathbf{B}^T \mathbf{R}^{-1}_{\mathbf{c}} \mathbf{B})^{-1}$. Thus between successive poses, proprioceptive odometery measurements are augmented with exteroceptive measurements.

A robot may close a loop and return to a previously visited location and re-observe features. At loop closure, we may solve Eq. \ref{eq:rot-constraint-rel-feats} to estimate the relative rotation between two poses $x_p$ and $x_q$. In this case, $\mathbf{\hat{C}}_{qp, init} = \mathbf{\hat{C}}_{p} \mathbf{\hat{C}}^T_{q}$, where $\mathbf{\hat{C}}_{p} ,\mathbf{\hat{C}}^T_{q}$ are estimated by chaining together successive relative rotation estimates computed according to Eq. \ref{eq:heading-linear-problem}.
Note that Eq. \ref{eq:rot-constraint-rel-feats} can be solved similar to Eq. \ref{eq:heading-linear-problem} to compute the relative rotation constraint at loop closure. Once Eq. \ref{eq:heading-linear-problem} (or Eq. \ref{eq:rot-constraint-rel-feats}) is solved, it needs to be ensured that the solution is an orthogonal rotation, thus we project it back onto the $\text{SO}(2)$ manifold as $\mathbf{\hat{c}}_{proj} = \boldsymbol{\eta}(\mathbf{\hat{c}})$, where $\boldsymbol{\eta}$ is a vector valued function (see Eq. \ref{eq:so2-projection}, Appendix \ref{appx:llsq-rotations}). The error covariance post projection is $\boldsymbol{\Sigma}_{\mathbf{c}_{proj}} = \bar{\nabla}\boldsymbol{\eta}|_{\mathbf{\hat{c}}} \Sigma_{\mathbf{c}} \bar{\nabla}^T\boldsymbol{\eta}|_{\mathbf{\hat{c}}}$ where $\bar{\nabla}\boldsymbol{\eta}|_{\mathbf{\hat{c}}}$ is the Jacobian of projection function $\boldsymbol{\eta}$ computed at the estimated values. Future references to $\mathbf{c}$ will drop the projection subscript for clarity.

Once $\mathbf{\hat{c}}$ is computed, it is transformed into the relative heading angle value (Eq. \ref{eq:c-vec-to-angle}, Appendix Appendix \ref{appx:llsq-rotations}), which in 2D is the scalar $\widehat{\delta \theta}$. Planar SLAM has the property that relative orientation measurements are linear in heading by virtue of which we can formulate the following linear problem

\begin{equation}\label{eq:global-heading-linear-problem}
	\widehat{\boldsymbol{\delta \theta}} = \mathbf{H}\boldsymbol{\theta} + \mathbf{v}_{\theta},
\end{equation}

where $\widehat{\boldsymbol{\delta \theta}}$ is the vector all relative orientation measurements, $\mathbf{H}$ is a matrix composed of elements from the set $\{-1,0,+1\}$ and $\boldsymbol{\theta}$ is the vector of robot heading angles. However, solving Eq. \ref{eq:global-heading-linear-problem} directly may not provide the correct answer as the linear least squares formulation is indifferent to the angle wrap-around problem. In the proceeding section we describe how to overcome this problem.
Lastly, we may compute the information matrix of the global heading estimate from Eq. \ref{eq:global-heading-linear-problem} as 
$\boldsymbol{\Omega}_{\boldsymbol{\theta}} = \mathbf{H}^T \mathbf{R}^{-1}_{\theta} \mathbf{H}$ where $\mathbf{R}_{\theta}$ is a diagonal matrix composed of uncertainty in relative orientation estimates. In Section \ref{subsec:global-estimation} we show how information matrix $\boldsymbol{\Omega}_{\boldsymbol{\theta}}$ is used by our algorithm to compute the map and history of robot positions.

\subsubsection{On-Manifold Optimization Using Relative Orientation Measurements}\label{subsec:heading-estimation-manopt}

The method described previously allows us to estimate relative rotations between poses. The set of poses and constraints from relative rotation estimates form a graph $\mathcal{G} = (\mathcal{V,E})$ whose nodes $\mathcal{V} = \{\nu_1,\dots,\nu_n\}$ are the pose orientations and whose edge $\epsilon_{pq} \in \mathcal{E}$ is a relative orientation constraint between nodes $\nu_p,\nu_q$. The problem at hand is to compute the global orientations for all nodes given relative rotation measurements.

Let $\mathbf{\hat{C}}_{qp}$ be the estimate of DCM $\mathbf{C}_{qp}$ for the relative rotation between nodes $\nu_p,\nu_q$. In the noise free measurement case, $\mathbf{\hat{C}}_{qp} \mathbf{C}_q = \mathbf{C}_p$. However, given a set of noisy measurements we minimize $ \sum_{\epsilon_{pq} \in  \mathcal{E}} \kappa_{qp} || \mathbf{\hat{C}}_{qp} \mathbf{C}_q - \mathbf{C}_p ||_{F}$ where $||\cdot||_{F}$ denotes the Frobenius matrix norm and $\kappa_{qp}$ is a weight for the measurement $\mathbf{\hat{C}}_{qp}$. Now the Frobenius norm can be expanded as follows,

\begin{equation}\label{eq:frob-norm-to-trace}
|| \mathbf{\hat{C}}_{qp} \mathbf{C}_q - \mathbf{C}_p ||^2 = ||\mathbf{C}_p||^2 + ||\mathbf{C}_q||^2 - 2\text{tr}(\mathbf{C}^T_q \mathbf{\hat{C}}^T_{qp}\mathbf{C}_p).
\end{equation}

Thus minimizing the Frobenius norm is equivalent to minimizing the term $-\text{tr}(\mathbf{C}^T_q \mathbf{\hat{C}}^T_{qp}\mathbf{C}_p)$ where $\text{tr}(\cdot)$ denotes the trace operator. Using properties of trace ($\text{tr}(\mathbf{X}) = \text{tr}(\mathbf{X}^T)$), we have the cost function to minimize as

\begin{equation}\label{eq:rotation-optimization}
J = -\sum_{\epsilon_{pq} \in  \mathcal{E} } \kappa_{qp} ~ \text{tr}(\mathbf{C}^T_{p} \mathbf{\hat{C}}_{qp}\mathbf{C}_{q}),
\end{equation}

where $\kappa_{qp} = 1 / \sigma_{\delta \theta_{qp}}$, i.e., inverse of standard deviation of relative rotation estimate. The Euclidean gradients for the cost function $J$ are

\begin{equation}\label{eq:global-rot-gradient-m}
\frac{\partial J}{\partial \mathbf{C}_{p}}  = -\sum_{\epsilon_{pq} \in  \mathcal{E} } \kappa_{qp} \mathbf{\hat{C}}_{qp} \mathbf{C}_{q}, ~\frac{\partial \mathcal{F}_{c}}{\partial \mathbf{C}_{q}}  =-\sum_{\epsilon_{pq} \in  \mathcal{E} } \kappa_{qp}\mathbf{\hat{C}}^T_{qp} \mathbf{C}_{p}.
\end{equation}


%
Note that in the cost function given by Eq. \ref{eq:rotation-optimization}, we directly optimize over the set of orientations for all poses. The initial guess can be computed by chaining together relative rotation estimates computed in Section \ref{subsec:heading-estimation-llsq}. Another way of looking at Eq. \ref{eq:rotation-optimization} is as follows, we have $\mathbf{\hat{C}}_{qp} = \mathbf{V}_{qp} \mathbf{C}_p \mathbf{C}_q$ where $\mathbf{V}_{qp}$ is the perturbation due to noise. Then solving Eq. \ref{eq:rotation-optimization} is equivalent to computing the maximum likelihood estimator with a Langevin prior on the perturbation $\mathbf{V}_{qp}$ where $\kappa_{qp}$ becomes the Langevin concentration parameter \cite{boumal-cdc13}. We use the Manopt MATLAB toolbox developed in \cite{JMLR:v15:boumal14a} to minimize the cost function $J$ using trust regions based optimization routine \cite{AbsBakGal2007-FoCM}.

\subsection{Global Trajectory and Feature Estimation}\label{subsec:global-estimation}

Let ${^l}\hat{\boldsymbol{\Delta}} \sim \mathcal{N}({^l}\boldsymbol{\Delta}, {^l}\mathbf{R}_{\boldsymbol{\Delta}}=blkdiag([{^l}\mathbf{R}_{\boldsymbol{\Delta}_1},{^l}\mathbf{R}_{\boldsymbol{\Delta}_2} \dots]))$ be the vector of all local relative position measurements from robot to features. After computing the global orientations according to Section \ref{subsec:heading-estimation-manopt}, the vector of local relative measurements ${^l}\hat{\boldsymbol{\Delta}}$ can be transformed to the world frame similar to Eq. \ref{eq:local2global-positions}.
From the transformed global measurements we can formulate the linear estimation problem as

\begin{equation}\label{eq:mapping-llsq-uncorrelated}
{^w}\hat{\boldsymbol{\Delta}} =  \mathbf{\hat{C}}^T ~ {^l}\hat{\boldsymbol{\Delta}}= \mathbf{A'} 
\begin{bmatrix}
\mathbf{p} \\
\mathbf{l}
\end{bmatrix} 
+ {^w}\mathbf{v}_{\boldsymbol{\Delta}},
\end{equation}

where $\mathbf{\hat{C}} = \mathbf{C} (\boldsymbol{\hat{\theta}})$ is the corresponding composition of DCM matrices parametrized by the estimated heading $\boldsymbol{\hat{\theta}}$, $[\mathbf{p}^T ~ \mathbf{l}^T]^T$ is the vector of robot and feature positions, $\mathbf{A}'$ is a matrix with each row containing elements of the set $\{-1,0,+1\}$ and ${^w}\mathbf{v}_{\boldsymbol{\Delta}} \sim \mathcal{N}(\mathbf{0}, {^w}\mathbf{R}_{\boldsymbol{\Delta}} = \mathbf{C}^T{^l}\mathbf{R}_{\boldsymbol{\Delta}}\mathbf{C})$ is the noise vector. 
If we were to solve for the feature positions directly from Eq. \ref{eq:mapping-llsq-uncorrelated}, we would end up with an incorrect estimate as the global orientation estimates $\boldsymbol{\hat{\theta}}$ are correlated. Thus relative feature measurements in the global frame are correlated with heading estimates as well. We now describe how to setup the position estimation problem while correctly incorporating the appropriate error covariances similar to the trick employed in LAGO \cite{Carlone-LAGO}.
After computing the orientation estimates $\boldsymbol{\hat{\theta}}$ along with the transformed global relative robot to feature measurements we stack them to give us a new measurement vector $\boldsymbol{\gamma}$. Then we have

\begin{equation}\label{eq:h_w}
\boldsymbol{\gamma} = 
\mathbf{h}_{w}({^l}\boldsymbol{\Delta},\boldsymbol{\theta}) + \mathbf{v}_w
 =  
\begin{bmatrix}
\mathbf{\hat{C}}^T ~ {^l}\hat{\boldsymbol{\Delta}}\\
\boldsymbol{\hat{\theta}}
\end{bmatrix} =
\underbrace{\begin{bmatrix}
	\mathbf{A}' & \mathbf{0} \\
	\mathbf{0} & \mathbf{I}
	\end{bmatrix}}_{\mathbf{A}}
\begin{bmatrix}
\mathbf{p} \\
\mathbf{l}\\
\boldsymbol{\theta}
\end{bmatrix}+
\begin{bmatrix}
{^w}\mathbf{v}_{\boldsymbol{\Delta}}\\
\mathbf{v}_{\boldsymbol{\theta}}
\end{bmatrix}.
\end{equation}

The error covariance $\mathbf{R}_{\boldsymbol{\gamma}}$ of measurement vector $\boldsymbol{\gamma}$ is then given by,

\begin{equation}
\mathbf{R}_{\boldsymbol{\gamma}} = \bar{\nabla}\mathbf{h}_w
\begin{bmatrix}
{^l}\mathbf{R}_{\boldsymbol{\Delta}} & \mathbf{0} \\
\mathbf{0} & \boldsymbol{\Sigma}_{\boldsymbol{\theta}} \\
\end{bmatrix}
\bar{\nabla}^T\mathbf{h}_w
\end{equation}

where $\bar{\nabla}\mathbf{h}_w$ is the Jacobian of measurement function $\mathbf{h}_w$ (Eq. \ref{eq:h_w}) given by

\begin{equation}
\bar{\nabla}\mathbf{h}_w =
\begin{bmatrix}
 \mathbf{C}^T & \mathbf{M}~ {^l}\hat{\boldsymbol{\Delta}} \\
\mathbf{0} & \mathbf{I}\\
\end{bmatrix},
\end{equation}

where $\mathbf{M} = \frac{\partial \mathbf{C}^T}{\partial \boldsymbol{\theta}}$. Thus we have

\begin{equation}
\mathbf{R}_{\boldsymbol{\gamma}} = 
\begin{bmatrix}
{^w}\mathbf{R}_{\boldsymbol{\Delta}} + \mathbf{M}\boldsymbol{\Sigma}_{\boldsymbol{\theta}}\mathbf{M}^T &  \mathbf{M}\boldsymbol{\Sigma}_{\boldsymbol{\theta}}\\
\boldsymbol{\Sigma}_{\boldsymbol{\theta}}\mathbf{M}^T & \boldsymbol{\Sigma}_{\boldsymbol{\theta}}
\end{bmatrix}.
\end{equation}

Finally, the solution to the linear estimation problem of Eq. \ref{eq:h_w} is given by 

\begin{equation}\label{eq:llsq-traj-feats}
\begin{bmatrix}
\mathbf{p}^* \\
\mathbf{l}^* \\
\boldsymbol{\theta}^*
\end{bmatrix} = (\mathbf{A}^T \mathbf{R}^{-1}_{\boldsymbol{\gamma}} \mathbf{A})^{-1} \mathbf{A}^T \mathbf{R}^{-1}_{\boldsymbol{\gamma}} \boldsymbol{\gamma}.
\end{equation}

Note that Eq. \ref{eq:llsq-traj-feats} involves the inversion of a large sparse matrix $\mathbf{R}_{\boldsymbol{\gamma}}$ which may not be suitable for implementation due to complexity and potential numerical issues. However, this inversion is easily avoided by analytically computing the information matrix $\boldsymbol{\Omega}_{\boldsymbol{\gamma}} = \mathbf{R}^{-1}_{\boldsymbol{\gamma}}$ using block-matrix inversion rules as

\begin{equation}
\boldsymbol{\Omega}_{\boldsymbol{\gamma}} = 
\begin{bmatrix}
{^w}\mathbf{R}^{-1}_{\boldsymbol{\Delta}} &  -{^w}\mathbf{R}^{-1}_{\boldsymbol{\Delta}}\mathbf{M} \\
-\mathbf{M}^T {^w}\mathbf{R}^{-1}_{\boldsymbol{\Delta}} & \boldsymbol{\Omega}_{\boldsymbol{\theta}} + \mathbf{M}^{T} {^w}\mathbf{R}^{-1}_{\boldsymbol{\Delta}}\mathbf{M}
\end{bmatrix}.
\end{equation}

\subsection{Extending RFM-SLAM to 3D}\label{subsec:method-remarks}

The global orientation optimization problem given relative measurements (Eq. \ref{eq:rotation-optimization}) does not change from 2D to 3D. A minor difference arises in solving for relative orientation at loop closure (Eq. \ref{eq:rot-constraint-rel-feats}) where a robot would require observations to 3 features from two poses as 9 constraints are required to solve for the DCM ($\mathbf{C}_k \in \mathbb{R}^{3 \times 3}$ in SO(3)). Further, the linear position estimation problem of Eq. \ref{eq:h_w} also remains identical. The key difference occurs in computing the uncertainty over global orientation estimates as the 3D rotation problem cannot be setup similar to the 2D case (Eq. \ref{eq:global-heading-linear-problem}). In 3D, relative orientations measurements are not linear in robot orientation, rather they are non-linear functions of rotation parameters. In this regard, the work of \cite{dorst2005first} develops an analysis for first-order error propagation in 3D rotation estimation which may be applicable to 3D RFM-SLAM. Investigating this aspect of the estimation problem forms part of our future work.

\section{Results}\label{sec:plum1-results}
\begin{figure}[ht!]
	\centering
	\subfigure[Map M1 with $1129$ robot poses and $286$ landmarks. The robot trajectory is $544.50$m long with 2 loop closures but robot does not return to start.]{\includegraphics[width=1.5in]{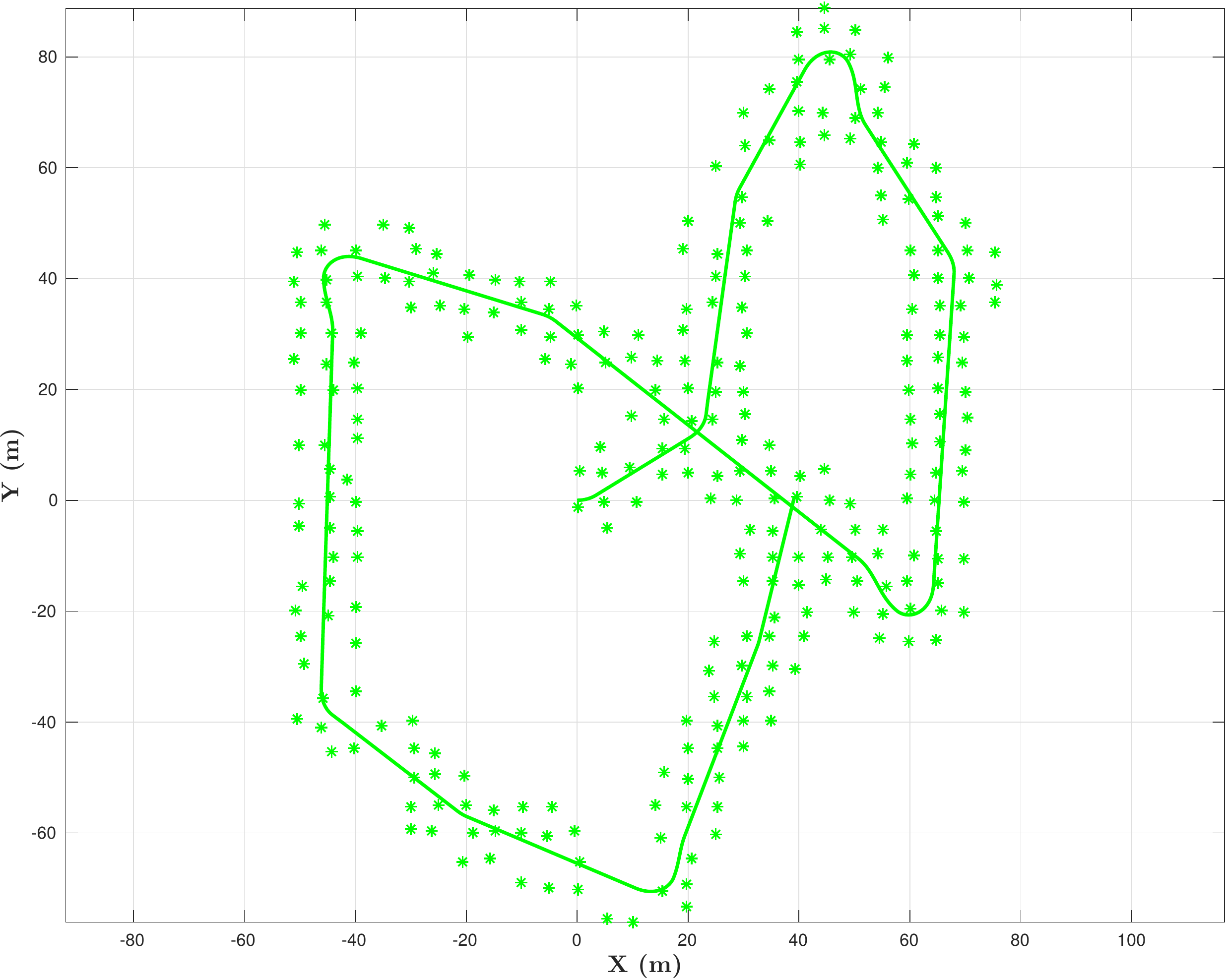}\label{subfig:mapM1}}
	\hspace{0.1in}
	\subfigure[Map M2 with $2064$ robot poses and $777$ landmarks. The robot trajectory is $1000.87$m long with 5 loop closures.]{\includegraphics[width=1.5in]{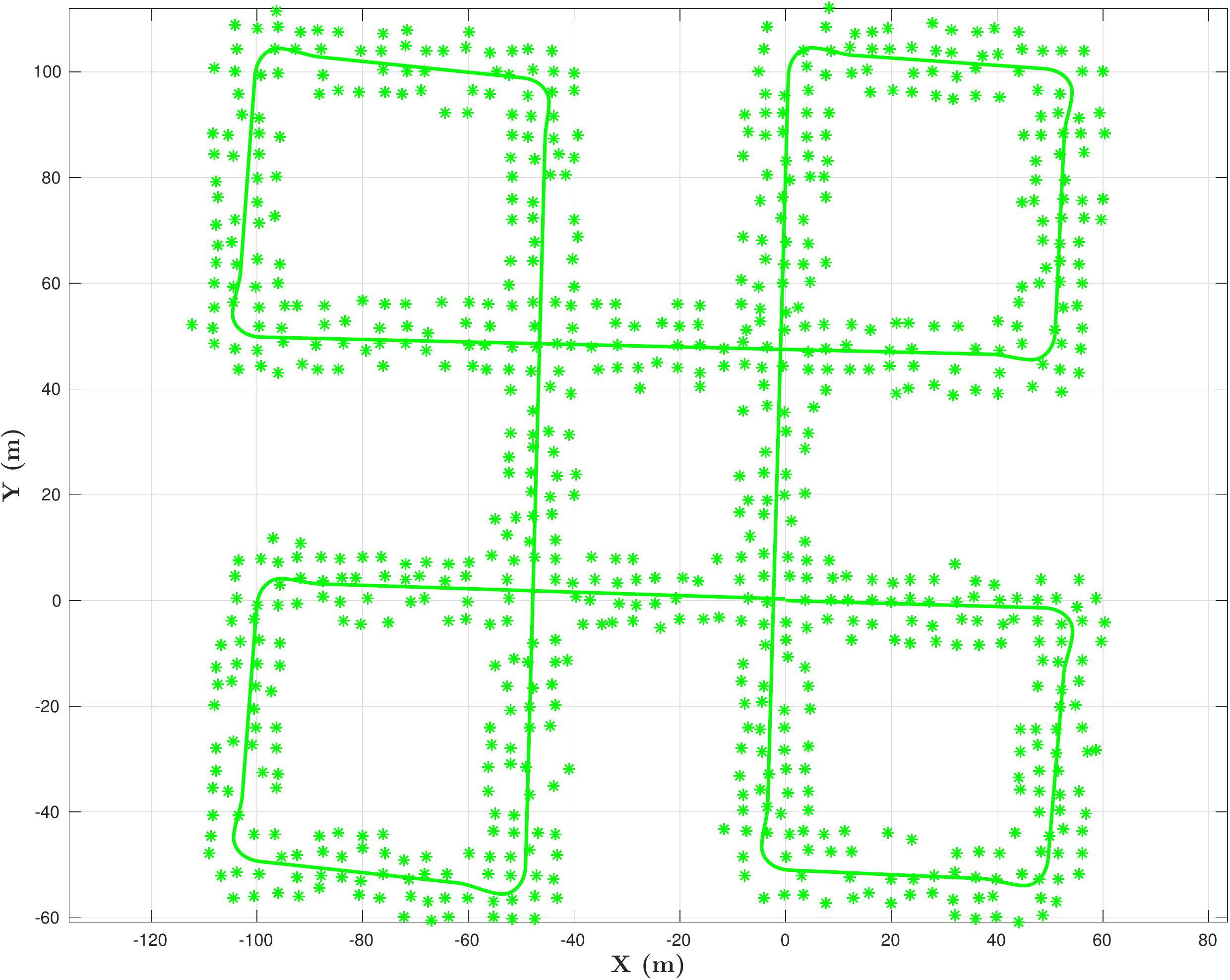}\label{subfig:mapM2}}
	\caption{The two scenarios used in the simulations and comparisons.}
	\label{fig:environments}	
\end{figure}

We conducted $1600$ simulations in total for two planar maps M1 and M2 (see Figs. \ref{subfig:mapM1} and \ref{subfig:mapM2}). The maps themselves were constructed by randomly sampling landmarks in a 2D environment after which simulated sensor data was collected by driving the robot around a sequence of waypoints. Proprioceptive odometery noise $\sigma_{odo}$ is varied by scale factor $ \alpha = \{ 1,2,3,4 \}$, where $\alpha =1 $ corresponds to $\sigma_{odo} = diag([0.05\text{m}, 0.05\text{m},0.6^{\circ}])$ and range bearing sensor noise $\sigma_{rb}$ is varied by scale factor $ \beta = \{ 1,2,3,4 \}$, where $\beta =1 $ corresponds to $\sigma_{rb}=diag([0.05\text{m}, 0.6^{\circ}])$. For each map, $50$ simulations were conducted for each fixed noise level and $16$ variations of noise values were used in total. For each simulation the resulting data was processed by both RFM-SLAM (MATLAB) and GTSAM (C++) \cite{dellaert2012factor}. GTSAM utilized the Levenberg-Marquardt Algorithm and both Manopt \cite{JMLR:v15:boumal14a} and GTSAM were allowed a maximum of 100 iterations.
We now proceed to discuss our results in the context of key aspects that affect solution accuracy, i.e., map, odometery noise and range bearing sensor accuracy.

\subsection{Changing Map}

Figure \ref{fig:environments} shows the two maps; map M1 with $1129$ robot poses and $286$ landmarks; and map M2 with $2064$ robot poses and $777$ features. Each map presents a different challenge, i.e., in M1 there are 2 loop closures and robot trajectory does not terminate at the start location, whereas in M2 there are 5 loop closures and robot returns to its start location.
Table \ref{table:rmse-50runs-rfmslam} shows that GTSAM average RMSE in robot pose is greater for map M2 than M1 for all noise combinations except for $\alpha=4, \beta = 1$. We note GTSAM suffers more catastrophic failures in map M2 than map M1 (Table \ref{table:rmse-50runs-rfmslam}). This is despite the fact that there are more loop closures in M2 and robot returns to start. The previous observation may be attributed to the trajectory in M2 ($\approx 1000$m) being longer than in M1 ($\approx 500$m) which results in odometery based initial guess being further from the ground truth than for map M1. An interesting difference emerges, for all noise combinations in the case of RFM-SLAM, average RMSE for map M2 is smaller than that for M1 despite the trajectory in M2 being twice as long as that of M1. This may be attributed to two factors; RFM-SLAM is able to exploit the graph topology for M2 (multiple loop closures) in the orientation estimation phase\footnote{An excellent insight into the problem of how graph topology affects SLAM accuracy is provided in \cite{khosoussi-iros14, khosoussi-icra16}.}; using range bearing measurements to augment relative orientation estimation provides a measure of robustness to the on-manifold optimization problem and purely odometery-based initial guess plays no role in the estimation process.

\begin{figure}
	\centering
	\subfigure[RMSE vs. $\beta$ for map M1.]{\includegraphics[width=1.4in]{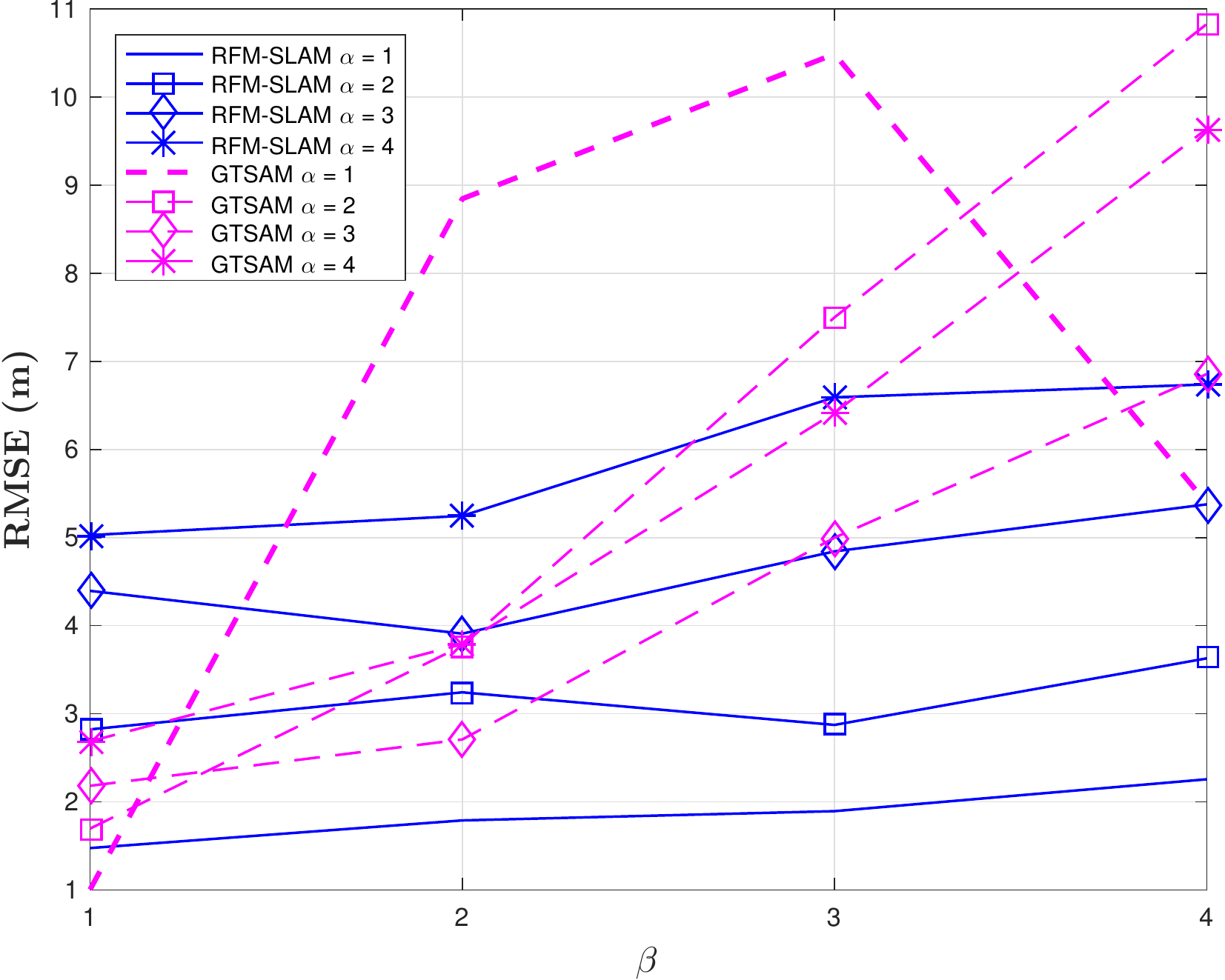}\label{subfig:mapl1-rmse-vs-beta}}	
	\subfigure[RMSE vs. $\beta$ for map M2.]{\includegraphics[width=1.4in]{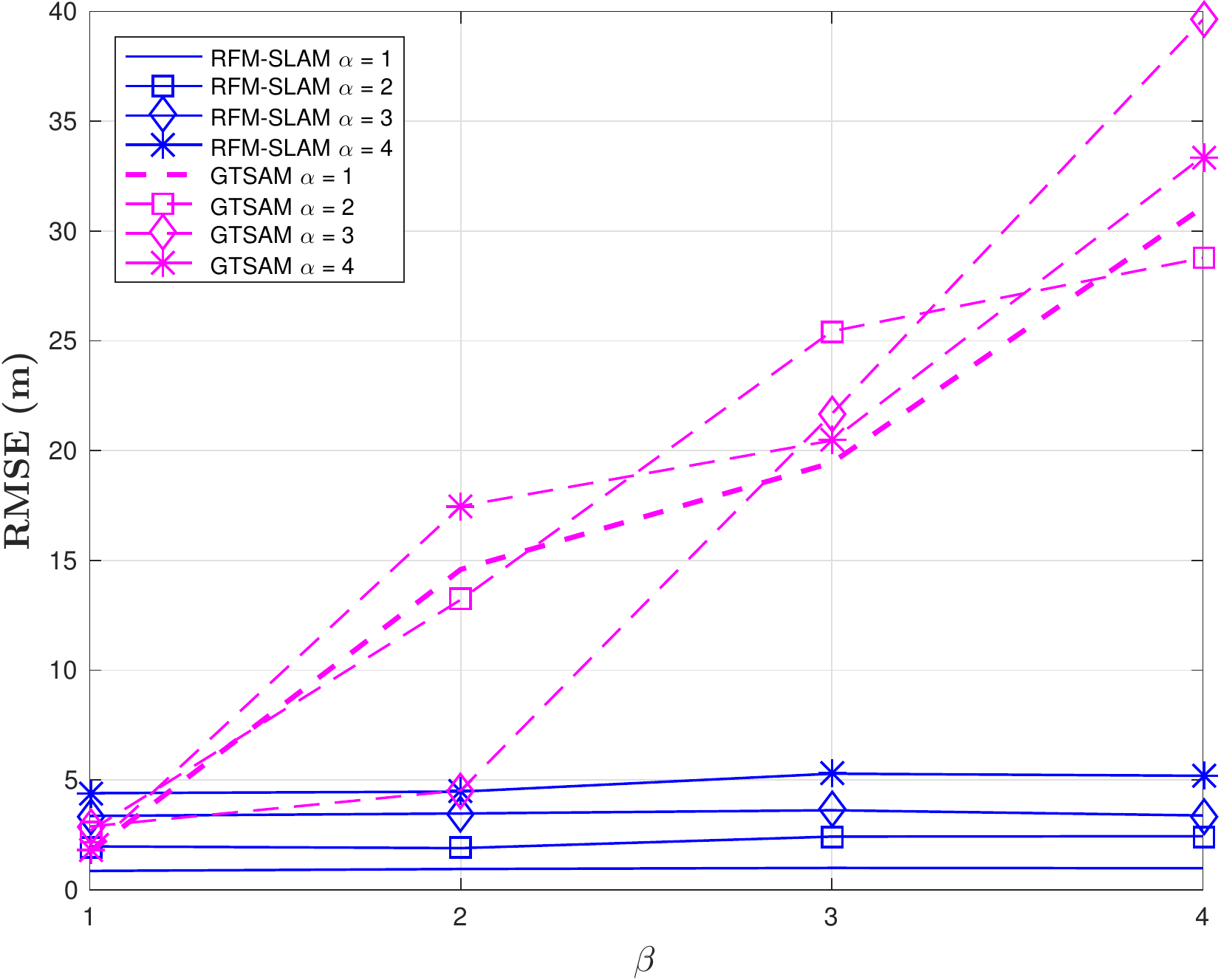}\label{subfig:maps4-rmse-vs-beta}}	
	\caption{Behavior of RMSE in robot position as odometery noise level $\beta$ is increased for different $\alpha$. The solid blue curves depict \txcb{RFM-SLAM} behavior and dashed magenta curves are for \txcm{GTSAM}.}
	\label{fig:RMSE-vs-beta}
\end{figure}

\begin{figure}[ht!]
	\centering
	\subfigure[RMSE vs. $\alpha$ for map M1.]{\includegraphics[width=1.5in]{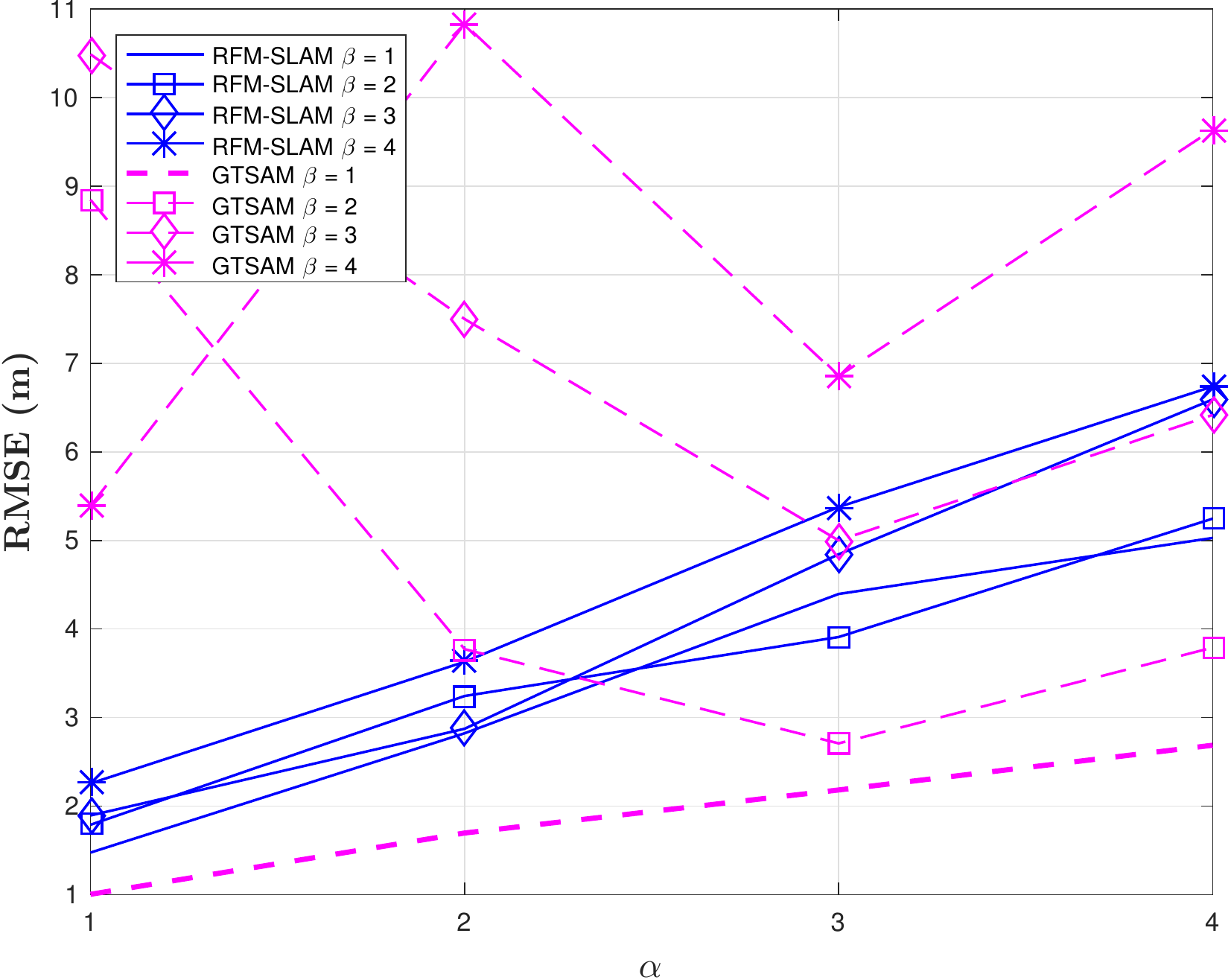}\label{subfig:mapl1-rmse-vs-alpha}}
	\hspace{0.1in}
	\subfigure[RMSE vs. $\alpha$ for map M2.]{\includegraphics[width=1.5in]{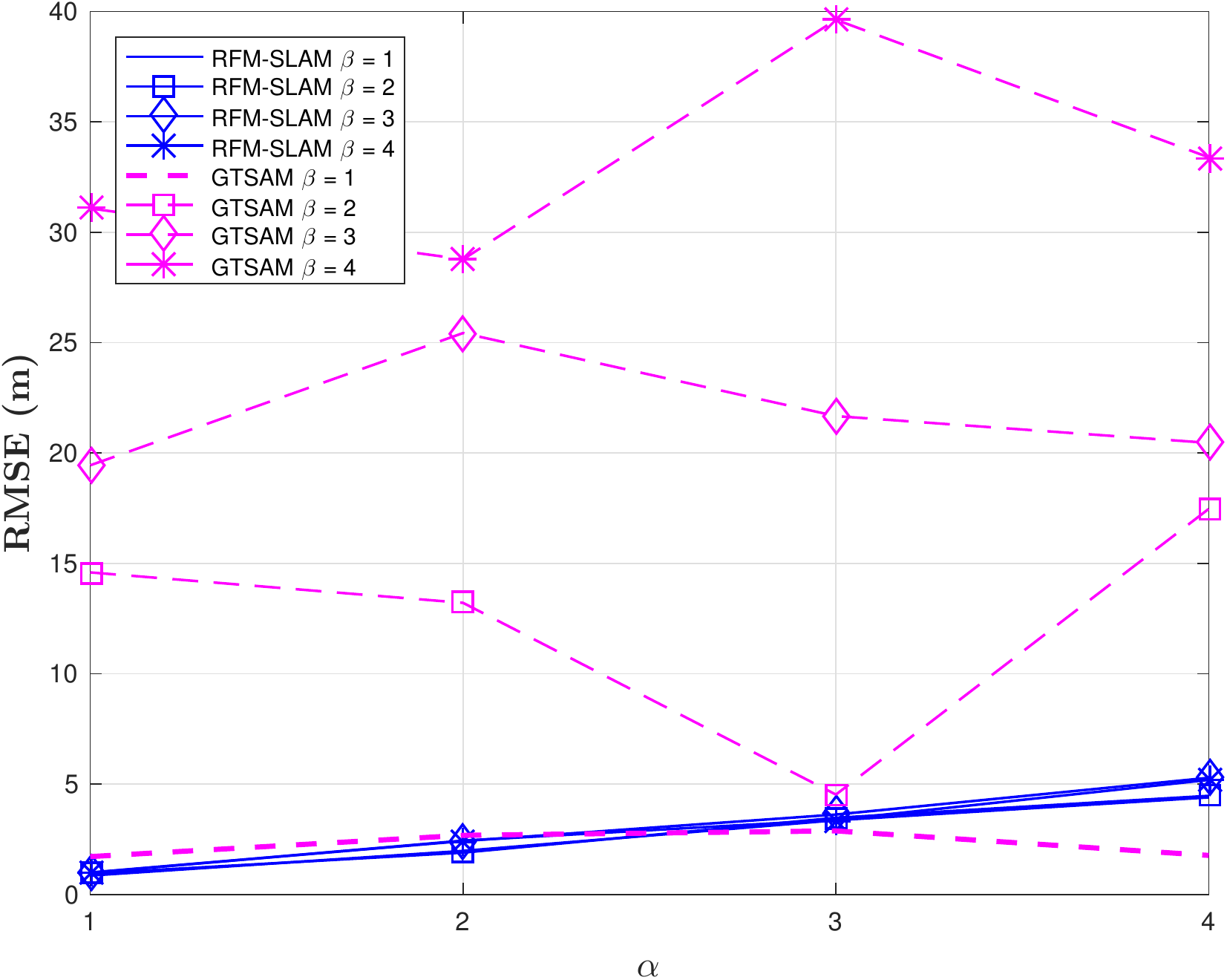}\label{subfig:maps4-rmse-vs-alpha}}	
	\caption{Behavior of RMSE in robot position as range bearing noise level $\alpha$ is increased for different $\beta$.}
	\label{fig:RMSE-vs-alpha}
\end{figure}

\subsection{Increasing Proprioceptive Odometery Noise}

Figure \ref{fig:RMSE-vs-beta} shows that for low odometery noise $\beta=1$ both methods perform comparably (same order of magnitude in RMSE) in both scenarios. For $\beta=1$, in the case of map M1 GTSAM performs slightly better than RFM-SLAM. Increasing the proprioceptive odometery noise has the effect of reducing the quality of initial guess that GTSAM relies on which is evident from Figs. \ref{subfig:mapl1-rmse-vs-beta} and \ref{subfig:maps4-rmse-vs-beta}. In both maps, as $\beta$ is increased, RFM-SLAM performance degrades much slower compared to GTSAM, where in map M2 particularly (Fig. \ref{subfig:maps4-rmse-vs-beta}) GTSAM shows a rapid decline in solution accuracy. We take the case of $\alpha=1$ to highlight the variation in solution accuracy as odometeric noise is increased from lowest ($\beta=1$) to its highest ($\beta=4$) value. In the case of M1, GTSAM solution accuracy degrades from $1.004$m to $5.389$m as the number of catastrophic failures increased from $0$ to $4$, whereas RFM-SLAM accuracy degrades from $1.475$m to $2.256$m. In the case of M2, GTSAM solution accuracy degrades rapidly by $1709.3$\% as RMSE rises from $1.718$m to $31.084$m due to the number of catastrophic failures rising from $1$ to $18$ whereas RFM-SLAM accuracy reduces gently from $0.859$m to $0.982$m. Thus simulation results show that RFM-SLAM solution accuracy degrades gracefully for both maps with increasing noise as it does not suffer catastrophic failure whereas GTSAM's performance is dominated by its sensitivity to the initial guess error (odometery).

\subsection{Increasing Range Bearing Sensor Noise}

Figure \ref{fig:RMSE-vs-alpha} shows that for the lowest odometeric noise value ($\beta=1$), both methods show a well defined behavior in RMSE growth as $\alpha$ increases. We look at the variation in error between lowest ($\alpha=1$) and highest ($\alpha=4$) range bearing sensor noise when proprioceptive odometery noise is lowest ($\beta = 1$). In map M1 as $\alpha$ increases from $1$ to $4$, RFM-SLAM RMSE rises from $1.475$m to $5.028$m, for GTSAM in the same map, we see a rise from $1.004$m to $2.687$m. In case of map M2, RFM-SLAM RMSE increases from $0.859$m to $4.4$m ($418$\% increase) whereas for GTSAM we see a rise from $1.718$m to $1.771$m. Thus RFM-SLAM exhibits a higher relative increase in RMSE than GTSAM for increasing $\alpha$. Thus simulation results show that compared to GTSAM, RFM-SLAM performance is dominated by range bearing sensor noise. 

\subsection{Discussion}\label{subsec:results-discussion}
Each method has a dominating factor that affects its behavior; for RFM-SLAM it is the range bearing sensor noise as we rely on this information in the orientation optimization phase; for GTSAM it is the proprioceptive odometery as it relies on odometery to bootstrap the solver. However, our results indicate that for all noise values, RFM-SLAM remains free of catastrophic failures due to which RMSE growth behaves well unlike in the case of GTSAM where the propensity of catastrophic failures increases with odometeric noise. In the case of GTSAM we see an order of magnitude increase in maximum RMSE over RFM-SLAM ($\approx 40$m vs. $\approx 7$m) at $\alpha=3,\beta = 4$. In few cases, GTSAM failed to converge to a solution, these numbers are also reported in Table \ref{table:rmse-50runs-rfmslam}. Further as the number of robot poses grows, odometery based initial guess diverges in an unbounded manner which may tend to dominate the solution accuracy in existing methods compared to noise in range bearing sensing. These results indicate a clear benefit of separating orientation and position estimation as it enhances robustness and reliability of the SLAM solution.

\begin{table*}[]
		\begin{center}
		\footnotesize
		\begin{tabular}{|p{1cm}|| c | c || c | c || c | c || c | c |}
			\hline
			\multirow{3}{1cm}{Scenario} & \multicolumn{8}{c|}{Average RMSE in robot position estimate  \textbf{(number of catastrophic failures, number of cases with no convergence)}} \\
			\cline{2-9} &  \multicolumn{2}{c||}{$\alpha = 1$} & \multicolumn{2}{c||}{$\alpha = 2$} & \multicolumn{2}{c||}{$\alpha=3$} & \multicolumn{2}{c|}{$\alpha=4$}\\
			
			\cline{2-9} & \txcb{RFM-SLAM} & \txcm{GTSAM} & \txcb{RFM-SLAM} & \txcm{GTSAM} & \txcb{RFM-SLAM} & \txcm{GTSAM} & \txcb{RFM-SLAM} & \txcm{GTSAM} \\						
			\hline
			
			\hline 
			\multirow{3}{1cm}{} & \multicolumn{8}{c|}{$\beta = 1$}  \\
			
			\hline  
			
			M1 & 1.475 \textbf{(0)} & 1.004 \textbf{(0)} & 2.822 \textbf{(0)} & 1.696 \textbf{(0)} & 4.395 \textbf{(0)} & 2.182 \textbf{(0)} & 5.028 \textbf{(0)} & 2.687 \textbf{(0)} \\			
			\hline
			
			M2 & 0.859 \textbf{(0)} & 1.718 \textbf{(1)} & 1.976 \textbf{(0)} & 2.682 \textbf{(2)} & 3.365 \textbf{(0)} & 2.882 \textbf{(1)} & 4.400 \textbf{(0)} & 1.771 \textbf{(0)} \\	
			\hline
			
			\hline 
			\multirow{3}{1cm}{} & \multicolumn{8}{c|}{$\beta = 2$}  \\
			
			\hline  
			
			M1 & 1.789 \textbf{(0)} & 8.846 \textbf{(8)} & 3.243 (0) & 3.772 \textbf{(2)} & 3.909 \textbf{(0)} & 2.708 \textbf{(0)} & 5.246 \textbf{(0)} & 3.790 \textbf{(0)} \\			
			\hline
			
			M2 & 0.947 \textbf{(0)} & 14.593 \textbf{(10)} & 1.898 \textbf{(0)} & 13.214 \textbf{(7)} & 3.475 \textbf{(0)} & 4.528 \textbf{(1)} & 4.471 \textbf{(0)} & 17.472 \textbf{(8,5)} \\	
			\hline
			
			\hline
			\multirow{3}{1cm}{} & \multicolumn{8}{c|}{$\beta = 3$}  \\
			
			\hline  
			
			M1 & 1.894 \textbf{(0)} & 10.489 \textbf{(8)} & 2.873 \textbf{(0)} & 7.500 \textbf{(7)} & 4.843 \textbf{(0)} & 4.999 \textbf{(3,1)} & 6.592 \textbf{(0)} & 6.417 \textbf{(3,3)} \\				
			\hline
			
			M2 & 0.997 \textbf{(0)} & 19.445 \textbf{(13)} & 2.422 \textbf{(0)} & 25.437 \textbf{(16,1)} & 3.623 \textbf{(0)} & 21.675 \textbf{(12,5)} & 5.288 \textbf{(0)} & 20.461 \textbf{(11,3)} \\	
			\hline
			
			\hline							
			\multirow{3}{1cm}{} & \multicolumn{8}{c|}{$\beta = 4$}  \\
			
			\hline  
			
			M1 & 2.256 \textbf{(0)} & 5.389 \textbf{(4)} & 3.629 \textbf{(0)} & 10.830 \textbf{(7)} & 5.377 \textbf{(0)} & 6.856 \textbf{(2,1)} & 6.738 \textbf{(0)} & 9.640 \textbf{(7,5)} \\			
			\hline
			
			M2 & 0.982 \textbf{(0)} & 31.084 \textbf{(18,1)} & 2.442 \textbf{(0)} & 28.777 \textbf{(15,1)} & 3.384 \textbf{(0)} & 39.631 \textbf{(21,4)} & 5.193 \textbf{(0)} & 33.397 \textbf{(17,4)} \\	
			\hline
		\end{tabular}
		\normalsize
		\caption{Average robot position RMSE in meters with the number of catastrophic failures and number of instances where there was no solution in bold brackets as odometery and range bearing sensor noise are varied. For each noise and map combination, 50 simulations were conducted and the RMSE in pose estimation was averaged over these simulations (excluding runs for which GTSAM did not converge to a solution).}
		\label{table:rmse-50runs-rfmslam}
	\end{center}	
\end{table*}

\section{Conclusions and Future Work}

In this work, a novel approach to solving the feature-based SLAM problem was presented that exploits separation of robot orientation from position estimation. Our proposed method RFM-SLAM undertakes a computationally cheaper optimization problem than standard graph-based approaches. Further, empirical results indicate that RFM-SLAM is able to avoid catastrophic failure and solution accuracy behaves well under varying noise conditions. We can safely conclude that decoupling orientation estimation from position exhibits a distinct advantage in that robust solutions can be obtained due to reduced risk of catastrophic failures. Future work involves implementing RFM-SLAM in more efficient frameworks, e.g. C++ to compare the time required to solve given problems with state-of-the-art solvers. Though the non-linear optimization problem for orientation may be susceptible to initial guess error, such an issue was not observed, perhaps the underlying nature of the orientation estimation problem is less sensitive to the initial guess. This is an interesting aspect of our approach which will be studied as part of future work. 

%

\bibliographystyle{plainnat}
\scriptsize
\bibliography{References}

\appendices

\section{Relative Measurements and Rotations in 2D}\label{appx:llsq-rotations}

\pbold{Parameterizing the Direction Cosine Matrix}Let the rotation from pose $x_p$ to $x_q$ be $\delta \theta$. The DCM $\mathbf{C}_{qp}$ for the relative rotation $\delta \theta$ between $x_p$ and $x_q$ is,

\begin{equation}\label{eq:2d-dcm-parameters}
\mathbf{C}_{qp} = \begin{bmatrix}
cos(\delta \theta) & -sin(\delta \theta) \\
sin(\delta \theta) & cos(\delta \theta)
\end{bmatrix}.
\end{equation}

Thus in planar scenarios the matrix $\mathbf{C}_{qp}$ is parameterized by the 2-vector $\mathbf{c}_{qp} = [cos(\delta \theta), sin(\delta \theta)]^T$. 

\pbold{Relative Feature Measurements-based Constraints on Orientation}Let a robot make observations to two landmarks $l_i$ and $l_j$ from poses $x_p$ and $x_q$ as shown in Fig. \ref{subfig:rel-orientation-2-poses}. Observing this pair of landmarks from both poses forms a relative orientation constraint $\mathbf{C}_{qp}$ between $x_p$ and $x_q$. Let ${^l}\mathbf{d}^{ij}_p$ and ${^l}\mathbf{d}^{ij}_q$ be the relative feature measurements made from $x_p$ and $x_q$ respectively, then we have the following relation ${^l}\mathbf{d}^{ij}_p = \mathbf{C}_{qp} {^l}\mathbf{d}^{ij}_q$.
Using Eq. \ref{eq:2d-dcm-parameters} in this relation and rearranging, we have the following constraint on the relative orientation parameters,

\begin{equation}\label{eq:det-rot-constraint}
\begin{bmatrix}
{^l}d^{ij}_{p,x}\\
{^l}d^{ij}_{p,y}
\end{bmatrix}
= 
\underbrace{
\begin{bmatrix}
{^l}d^{ij}_{q,x} & -{^l}d^{ij}_{q,y}\\
{^l}d^{ij}_{q,y} & {^l}d^{ij}_{q,x}
\end{bmatrix}}_{\mathbf{B}'_{qp}}
\begin{bmatrix}
cos(\delta \theta) \\
sin(\delta \theta)
\end{bmatrix}.
\end{equation}

\pbold{Projection onto SO(2) Manifold}As discussed in Section \ref{subsec:heading-estimation-llsq}, solving Eq. \ref{eq:heading-linear-problem} or Eq. \ref{eq:rot-constraint-rel-feats} does not provide an orthogonal rotation as the solution. Thus the linear least squares solution $\mathbf{\hat{c}}$ is projected back on the $\text{SO}2$ manifold by normalization

\begin{equation}
\mathbf{\hat{c}}_{normalized} = \boldsymbol{\eta}(\mathbf{\hat{c}}) =  \frac{\mathbf{\hat{c}}}{||\mathbf{\hat{c}}||}.
\end{equation}

Followed by computing the Jacobian 

\begin{equation}\label{eq:so2-projection}
\bar{\nabla}\boldsymbol{\eta} = \frac{1}{\sqrt{c^2_{1} + c^2_{2}}}\begin{bmatrix}
c^2_{2} & -c_{2}c_{1}\\
-c_{2}c_{1} & c^2_{1}
\end{bmatrix},
\end{equation}

and then transforming the covariance given by the linear problem as $\boldsymbol{\Sigma}_{\mathbf{c}_{normalized}} = \bar{\nabla}\boldsymbol{\eta} \Sigma_{\mathbf{c}} \bar{\nabla}^T\boldsymbol{\eta}$. We drop the normalized subscript for readability. From the projected DCM parameters we can compute the rotation angle

\begin{equation}\label{eq:c-vec-to-angle}
\delta \hat{\theta} = tan^{-1}(\frac{c_2}{c_1}).
\end{equation}

\end{document}